\documentclass[sigconf,authorversion,nonacm]{acmart}

\AtBeginDocument{%
  \providecommand\BibTeX{{%
    \normalfont B\kern-0.5em{\scshape i\kern-0.25em b}\kern-0.8em\TeX}}}

\usepackage{multicol}
\usepackage{multirow}
\usepackage{graphicx}
\usepackage{amsmath}
\usepackage{booktabs}
\usepackage{enumitem}
\usepackage{caption}
\usepackage{subcaption}

\DeclareMathOperator*{\argmin}{arg\,min}
\usepackage{xcolor}
\usepackage{pifont}

\usepackage{marvosym}
\usepackage{balance}
\newcommand{\modelname}{VQ-Trans}

\usepackage{xspace}

\makeatletter
\DeclareRobustCommand\onedot{\futurelet\@let@token\@onedot}
\def\@onedot{\ifx\@let@token.\else.\null\fi\xspace}

\def\eg{\emph{e.g}\onedot}

\def\etal{\emph{et al}\onedot}
\makeatother

\begin{document}

\title{TeViS: Translating Text Synopses to Video Storyboards}

\author{Xu Gu$^{\dagger}$}
\thanks{$^{\dagger}$\ Both authors contributed equally to this research.}
\email{guxu@ruc.edu.cn}
\orcid{0009-0003-4847-6593}
\affiliation{
  \institution{Renmin University of China}
  \city{Beijing}
  \country{China}
}

\author{Yuchong Sun$^{\dagger}$}
\email{ycsun@ruc.edu.cn}
\orcid{0009-0004-6559-5620}
\affiliation{
  \institution{Renmin University of China}
  \city{Beijing}
  \country{China}
}

\author{Feiyue Ni}
\email{nifeiyue@ruc.edu.cn}
\orcid{0009-0001-6062-3459}
\affiliation{
  \institution{Renmin University of China}
  \city{Beijing}
  \country{China}
}

\author{Shizhe Chen}
\email{shizhe.chen@inria.fr}
\orcid{0000-0002-7313-9703}
\affiliation{
  \institution{INRIA}
  \city{Paris}
  \country{France}
}

\author{Xihua Wang}
\email{xihuaw@ruc.edu.cn}
\orcid{0009-0002-9454-5775}
\affiliation{
  \institution{Renmin University of China}
  \city{Beijing}
  \country{China}
}

\author{Ruihua Song\textsuperscript{\Letter}}
\email{songruihua_bloon@outlook.com}
\thanks{\Letter\ Corresponding author.}
\orcid{0000-0002-2163-7401}
\affiliation{%
  \institution{Renmin University of China}
  \city{Beijing}
  \country{China}
}

\author{Boyuan Li}
\email{liboyuan@ruc.edu.cn}
\orcid{0009-0008-5487-9678}
\affiliation{%
  \institution{Renmin University of China}
  \city{Beijing}
  \country{China}
}

\author{Xiang Cao}
\email{xiangcao@acm.org}
\orcid{0000-0002-5142-2292}
\affiliation{
  \institution{Bilibili Inc.}
  \city{Beijing}
  \country{China}
}

\renewcommand{\authors}{Xu Gu, Yuchong Sun, Feiyue Ni, Shizhe Chen, Xihua Wang, Ruihua Song, Boyuan Li, Xiang Cao}
\renewcommand{\shortauthors}{Xu Gu et al.}

\begin{abstract}

A video storyboard is a roadmap for video creation which consists of shot-by-shot images to visualize key plots in a text synopsis.
Creating video storyboards, however, remains challenging which not only requires cross-modal association between high-level texts and images but also demands long-term reasoning to make transitions smooth across shots.
In this paper, we propose a new task called \textbf{Te}xt synopsis to \textbf{Vi}deo \textbf{S}toryboard (\textbf{TeViS}) which aims to retrieve an ordered sequence of images as the video storyboard to visualize the text synopsis.
We construct a \textbf{MovieNet-TeViS} dataset based on the public MovieNet dataset \cite{huang2020movienet}. It contains 10K text synopses each paired with keyframes manually selected from corresponding movies by considering both relevance and cinematic coherence.
To benchmark the task, we present strong CLIP-based baselines and a novel \modelname~model. \modelname~ first encodes text synopsis and images into a joint embedding space and uses vector quantization (VQ) to improve the visual representation. Then, it auto-regressively generates a sequence of visual features for retrieval and ordering.
Experimental results demonstrate that \modelname~significantly outperforms prior methods and the CLIP-based baselines. 
Nevertheless, there is still a large gap compared to human performance suggesting room for promising future work.
The code and data are available at: \url{https://ruc-aimind.github.io/projects/TeViS/}
\end{abstract}




\begin{teaserfigure}
    \centering
    \begin{subfigure}[b]{0.49\textwidth}
         \centering
         \includegraphics[width=\textwidth]{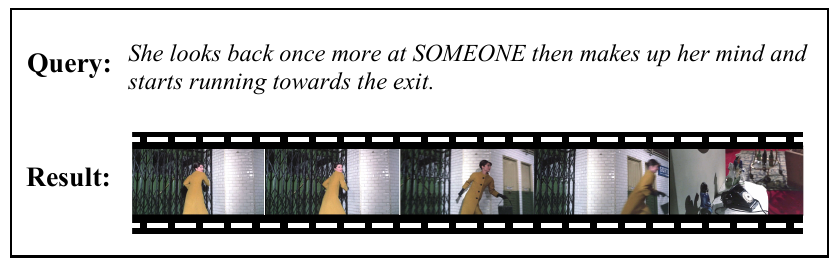}
         \caption{The text-to-video retrieval task (example from the LSMDC dataset \cite{rohrbach2017lsmdc}). The query text is detailed, and the candidate video is short containing redundant image frames.}
         \label{fig:lsmdc}
     \end{subfigure}
     \hfill
     \begin{subfigure}[b]{0.49\textwidth}
         \centering
         \includegraphics[width=\textwidth]{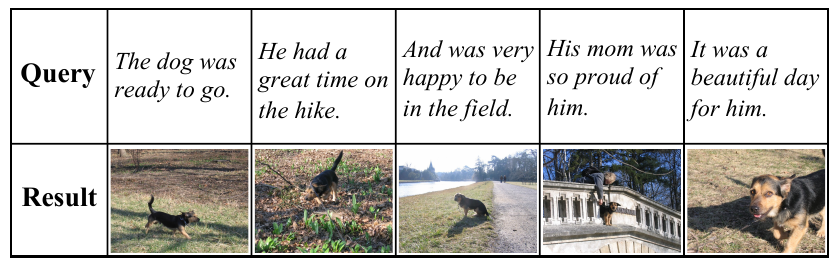}
         \caption{The story-to-image retrieval task (example from the VIST dataset \cite{huang2016vist}). Each sentence in the query text should be aligned with an image. 
         The transition in the image sequence is not coherent.
         }
         \label{fig:vist}
     \end{subfigure}
     \hfill
     \begin{subfigure}[b]{1\textwidth}
         \centering
         \includegraphics[width=\textwidth]{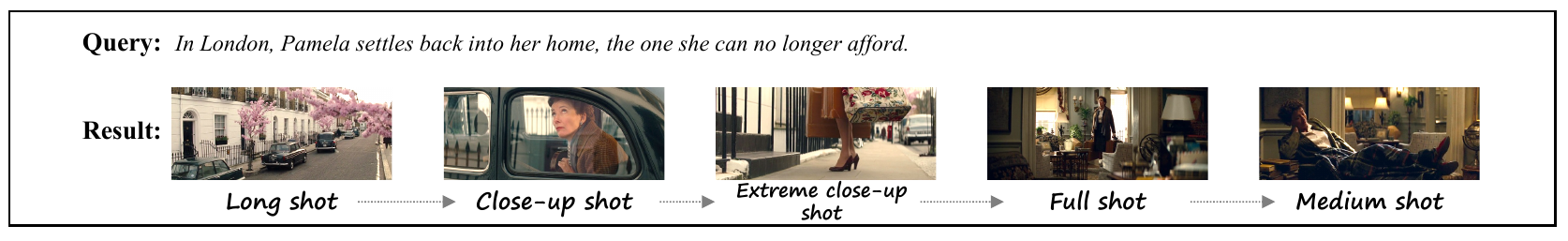}
         \caption{Our proposed text-to-storyboard task (example from our  MovieNet-TeViS dataset). It requires to retrieve a sequence of cinematically coherent images to visualize the succinct text synopsis. We manually construct the dataset from movies to preserve the language of the film's shot. }
         \label{fig:tevis}
     \end{subfigure}
\caption{Comparison of different cross-modal retrieval tasks. Our proposed text-to-storyboard task is more beneficial than existing tasks to assist amateurs in video creation.}
\label{fig:teaser}
\end{teaserfigure}

\maketitle

\section{Introduction}
With the prevalence of video sharing platforms, more and more video creators are emerging with enthusiasm to create videos using their own text synopses.
A critical step in professional video creation is to translate a text synopsis into a video storyboard\footnote{\url{https://creately.com/guides/how-to-make-a-storyboard-for-video/}}, which is a sequence of shot-by-shot images to visualize key plots in a screenplay.
Creating a high-quality video storyboard is however challenging for amateurs. It not only requires one to put relevant scenes, characters and actions in the video, but also demands for cinematic organizations of keyframes to enable coherent transitions across shots. 
Hence, there are high application needs in assisting amateurs to create more professional video storyboards from their text synopses. 

Although existing works have made great progress in text-to-video retrieval~\cite{lei2021clipbert,xue2021hdvila,miech2019howto100m,bain2021frozen}, 
story-to-image~\cite{hu2020visualStorytelling,Li2019storygan,chen2019storyboard} 
and even text-to-video generation~\cite{singer2022makeAvideo,ho2022imagenvideo,wu2022nuwa}, they are insufficient to support storyboard creation given the text synopsis.
The text-to-video retrieval or generation tasks mainly focus on short-term video clips with only a few seconds as shown in Fig.~\ref{fig:lsmdc}.
The images in these videos are highly redundant and cannot satisfy the requirement of a video storyboard for coherent keyframes.
The story-to-image work targets one-to-one mapping from $N$ sentences of detailed description to $N$ images (see Fig.~\ref{fig:vist}).
The transitions across images are not required to be smooth considering the lens language.
In addition, all these works care more about the text-image relevance with detailed query texts. 
The text synopses, however, are usually more abstract and brief as shown in Fig.~\ref{fig:tevis}. As a result, one-to-many mapping from the text to cinematically coherent images is necessary to visualize the text.
As shown in Fig.~\ref{fig:tevis}, to film the event of a woman returning home, a professional director first captures a distant frame showing the exterior of the house, then a close-up shot of the woman, an extreme close-up shot of the woman entering the house, and finally a full shot of the interior. 

In this paper, we propose a new task called \textbf{Te}xt synopsis to \textbf{Vi}deo \textbf{S}toryboard (\textbf{TeViS}).
In the TeViS task, we aim to retrieve an ordered sequence of images from a large-scale movie database as a video storyboard to visualize an input text synopsis.
For this purpose, we collect the \textbf{MovieNet-TeViS} dataset based on the public MovieNet dataset~\cite{huang2020movienet}. 
The MovieNet dataset contains high-level text synopses for movies and a coarse-grained alignment between movie segments and text synopses paragraphs.
We ask annotators to split paragraphs into semantically compact sentences and select a minimum set of keyframes from its aligned movie segment for each text synopsis sentence. Annotators should consider both relevancy to the text and cinematic coherency across frames for keyframe selection.
Finally, we obtain 10K text synopses, and each paired with 4.6 keyframes on average. 

We propose an auto-regressive generation model~\modelname~as a starting point to benchmark the TeViS task. 
~\modelname~ first encodes the text synopsis and images into a joint embedding space.
However, it is difficult to learn the language of cinematography from a vast and diverse collection of images.
Inspired by previous work using Vector Quantization to represent continuous data as discrete vectors ~\cite{van2017vqvae, esser2021vqgan, ramesh2021dalle}, 
we achieve this objective by combining recent advances in vector quantization and large language modeling.
Our~\modelname~uses vector quantization to improve the visual features and convert images from continuous visual features into discrete visual tokens.
Then, ~\modelname~auto-regressively generates a sequence of visual tokens where each token can be used to retrieve images from the candidate image pool.

\begin{table*}[t]
\centering
\caption{Comparison between MovieNet-TeViS and other movie datasets.}
\begin{tabular}{lcccccc}
\toprule
Datasets & avgDuration & avg\#Words & SecondsperWord & \#unique bi-grams & avgConcreteness \\
\midrule
LSMDC ~\cite{rohrbach2017lsmdc}  & 4.1s  & 9.03  & 0.4539 & 44.0K & 2.993     \\
MAD ~\cite{soldan2022mad}  & 4.04s & 12.69 & 0.3188 & 59.0K & 2.991     \\
CMD ~\cite{bain2020cmd} & 132s   & 18    & 7.33 & 83.2K & 2.598      \\ \midrule
MovieNet-TeViS (Ours)    & 63.7s    & 24.82 & 2.82 & 134.5K & 2.761       \\
\bottomrule
\end{tabular}
\label{tab:dataset}
\end{table*}

We design two settings to evaluate methods: i) an ordering setting that provides models with oracle keyframes to re-order conditioning on the text, and ii) a retrieving-and-ordering setting that requires models to retrieve relevant frames from 500 candidate images and order them. 
Experimental results show that employing Vector Quantization for image discretization and utilizing text as a prefix in a decoder-only model significantly improves the ordering performance.
Both quantitative and qualitative results show that our model is able to create reasonable video storyboards.

Our contributions are summarized as follows:
\parskip=0.1em
\begin{itemize}[itemsep=0.1em,parsep=0em,topsep=0em,partopsep=0em]
    \item We propose the \textbf{Te}xt Synopsis to \textbf{Vi}deo \textbf{S}toryboard task (\textbf{TeViS}) with the goal of retrieving an ordered sequence of images to visualize a high-level text synopsis.
    \item We construct a \textbf{MovieNet-TeViS} benchmark based on the MovieNet dataset~\cite{huang2020movienet}. It contains 10K text synopses with 4.6 keyframes on average for each synopsis.
    \item We establish a decoder-only baseline and propose Vector Quantization on frames to improve training stability and generation quality in long-term video storyboards. 
\end{itemize}
\section{Related Works}
Our work is related to previous works of two categories: text-to-vision and movie understanding.

\subsection{Text-to-Vision}
Text-to-vision aims to retrieve or generate visual information corresponding to an input text.
Inspired by the success of the pre-training paradigm in NLP~\cite{Devlin2018bert,brown2020gpt3}, recent advances in text-to-image retrieval also leverage massive image-text pairs to pre-train a large model for retrieval~\cite{Alec2021CLIP,huo2021wenlan,huang2021soho,chen2020uniter,xue2021probing,xue2023clipvip}. These methods achieve promising results on caption-based image retrieval tasks such as MSCOCO~\cite{chen2015mscoco}. 
CLIP~\cite{lei2021clipbert} adopts a dual-encoder architecture, uses 400 million image-text pairs for pre-training with a contrastive loss, and shows strong generalization power on cross-modal alignment.
Some works also pre-train video-language models on large-scale video-text pairs~\cite{xue2021hdvila,bain2021frozen,miech2019howto100m,ni2023showmevideo,sun2022LFVLP}.
However, text-to-image technologies can only produce static images which can not describe the dynamics in text synopsis, while text-to-video retrieval target searching existing video clips, rather than picking up keyframes from clips to collage out something new, which is demanded for if users input new texts.
Text-to-vision generation also develops rapidly. Earlier methods widely adopt GAN-based methods conditioned on text~\cite{reed2016generative,pan2017create,zhang2017stackgan}. Recent deep generative models use large Transformer networks~\cite{wu2022nuwa,ding2021cogview,ramesh2021dalle} or diffusion models~\cite{ramesh2022dalle2,saharia2022imagen} that can generate high-quality images. Text-to-video generation has been explored recently by extending advanced text-to-image generation methods~\cite{ho2022imagenvideo,singer2022makeAvideo}.  
However, even advanced text-to-video generation methods can only generate GIF-like short videos without complicated motions and dynamics. 

\subsection{Movie Understanding}

Existing works on movie understanding mainly explore content recognition and cinematic style analysis. Content recognition includes action~\cite{laptev2008movieaction,Bojanowski2013movieaction}, scene recognition~\cite{chasanis2008scene,han2011videoscene,Rao2020MovieScene}, and text-to-video retrieval task that focuses on movie datasets~\cite{rohrbach2017lsmdc,bain2020cmd}. Some works aim to analyze shot styles~\cite{rao2020shotType,Wang2009FilmShot}, movie genre~\cite{zhou2010moviegenre,simoes2016moviegenre} from a professional perspective.
There are also some movie-related datasets~\cite{rohrbach2017lsmdc,bain2020cmd,huang2020movienet}. 
LSMDC~\cite{rohrbach2017lsmdc} dataset contains short clips paired with human-annotated captions.
Condensed Movie Dataset (CMD) ~\cite{bain2020cmd} consists of key scenes from the movie, each of which is accompanied by a high-level semantic description of the scene.
The MAD~\cite{soldan2022mad} dataset is based on the LSMDC~\cite{rohrbach2017lsmdc} dataset. 
Our constructed dataset is built on MovieNet~\cite{huang2020movienet} dataset, which is a large collection of movies annotated with many kinds of tasks such as scene segmentation, cinematic style classification, story understanding and so on.

In terms of the text inputs, previous works have collected aligned script~\cite{zhu2022script}, caption~\cite{rohrbach2017lsmdc}, Descriptive Video Service (DVS)~\cite{torabi2015dvs}, book~\cite{Zhu2015book}, or synopsis~\cite{sun2022synopses} to movies. 
However, books cannot be well-aligned with adapted movies; DVS is hard to obtain and thus limited in scale; 
while scripts and captions are too detailed to compose for most non-professional users. We consider synopses are the most appropriate source which mimic the texts written by users in real scenario and contain desired level of details. MovieNet~\cite{huang2020movienet} and CMD~\cite{bain2020cmd} provide synopses but MovieNet is more appropriate as we explain in Sec.~\ref{sec:why}.

\section{MovieNet-TeViS Dataset}

In this section, we explain why we choose MovieNet as the basis, describe the dataset annotation process and present analyses on our dataset.

\subsection{Why MovieNet}
\label{sec:why}
Our goal is to assist amateur video makers to create video storyboards from text inputs. 
Since it is hard to obtain original video storyboards from professional video makers, we decided to select keyframes from released movies to reconstruct a succinct storyboard that a human user can use as a shooting plan.
Tab.~\ref{tab:dataset} presents the comparison of our dataset and related movie datasets. It shows that the duration of movie clips corresponding to a description in LSMDC and MAD is only 4 seconds and the average number of words in a description is only 9-12, which is much lower than ours. It is impossible to extract a meaningful storyboard from such short clips. Our MovieNet-TeViS and CMD give a synopsis or summary of 64-second or 132-second video segments respectively and thus we can expect such text is higher-level. Compared to CMD, our MovieNet-TeViS uses more words to describe video segments with half the duration of CMD. This indicates that our text synopses provide more details than those in CMD. As a start of such a challenging new task, our dataset is the most appropriate in duration of video clip and semantic level of text.

\begin{figure}[!t]
\centering
\includegraphics[width=0.85\linewidth]{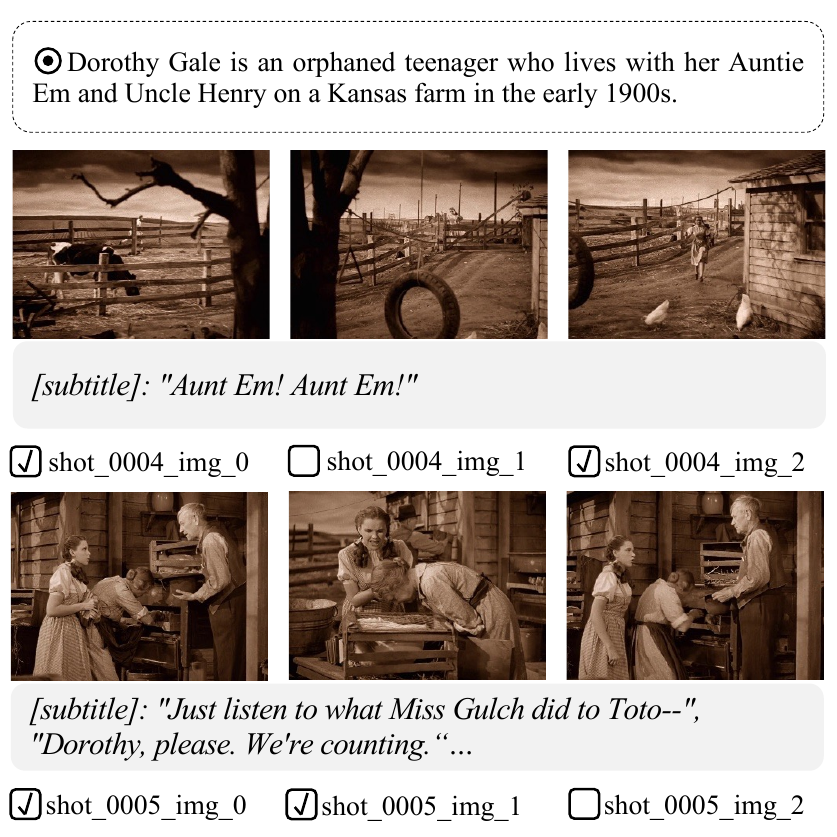}
\caption{Annotating keyframes of a storyboard for a text synopsis.}
\label{fig:anno_tool}
\end{figure}

\subsection{Data Annotation}
\label{sec:dataset_annotation}

MovieNet provides 4,208 text synopsis paragraph and movie segment pairs.
A paragraph consists of 8 sentences (113 words) on average and a segmentation contains 95 shots. 
It might be too difficult to learn semantic association and long-term reasoning over such long sequences with large variance. Therefore, we first split a paragraph into sentences and then align each sentence with a minimum number of keyframes in the movie segment.

\begin{figure}[htp]
\centering
\includegraphics[width=0.95\linewidth]{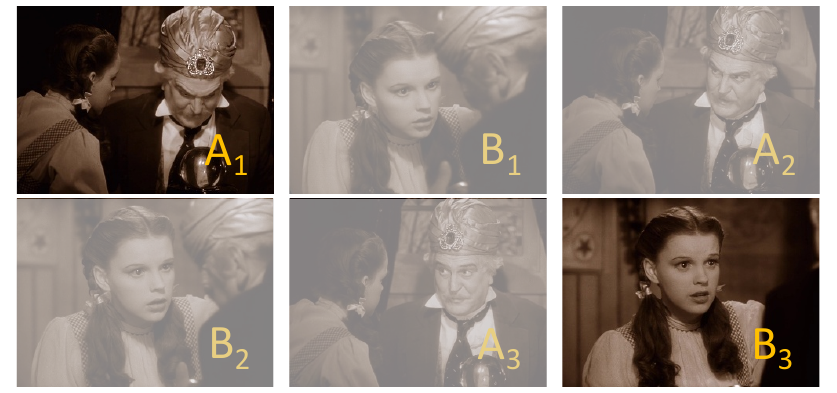}
\caption{Simplifying a storyboard by deleting redundant images.}
\label{fig:talk}
\end{figure}

Fig.~\ref{fig:anno_tool} shows the annotation interface. We present a synopsis paragraph sentence by sentence and all the shots aligned with the paragraph in MovieNet. For each shot, we display three evenly spaced frames as well as corresponding subtitles below the shot to help annotators understand the images better.
The annotator should first select the sentences to form a text synopsis, and then choose a minimum number of images to visualize the text.
We construct detailed guidelines to assure the quality of each annotated storyboard as follows:

\parskip=0.1em
\begin{enumerate}
[itemsep=0.1em,parsep=0em,topsep=0em,partopsep=0em]
    \item The number of keyframes should be less than 20. Split the sentence if the number of selected keyframes is more than 20, or filter the sentence out of the dataset;
    \item Do not select adjacent similar images, \eg, for the example in Fig.~\ref{fig:anno_tool},  shot\_0004\_img\_1 should be deleted when given shot\_0004\_img\_0;
    \item The image must add value to express the synopsis sentence in terms of relevancy or coherency, \eg, shot\_0005\_img\_2 cannot add any new value given shot\_0005\_img\_0;
    \item If there is a basic conversation with a cycle of repeated images, only keep one pattern to make the storyboard succinct by selecting the first images in the first cycle and the last image in the last cycle. 
    For example, in Fig.~\ref{fig:talk}, $A_{i} B_{i}$ is a basic conversation pattern and it has been repeated 3 times. We ask annotators to select $A_{1}$ and $B_{3}$ to compose a complete conversation. 
\end{enumerate}

To ensure data quality, our data are annotated in three rounds. We hire 60 annotators in the first round to select keyframes that are necessary for relevancy or a vision language; in the second round the annotators further simplify or revise the storyboards by consistent rules; and six volunteer experts in the third round review and finalize the selected keyframes. 

In addition, we find some non-adjacent images are still similar and some images in a conversation can be re-ordered without hurting the storyboard. Therefore, we further ask annotators to detect such interchangeable images over our testing data and add these re-ordered storyboards into the ground-truth set.

\subsection{Dataset Analysis\label{sec:dataset_analysis}}

\noindent\textbf{Dataset statistics.} 
Our collected MovieNet-TeViS dataset uses 2,949 paragraph-segment pairs from MovieNet after filtering improper examples by annotators. We sort storyboards by the number of keyframes ascendingly and use the first 10,000 pairs of a synopsis sentence in English and a video storyboard, i.e., a sequence of keyframes as our final dataset. 
There are 45,584 keyframes in total. 
The number of keyframes in a storyboard ranges from 3 to 11, and about 60\% of storyboards consist of 3 or 4 keyframes.
The average number of words in a synopsis sentence is about 24. 
In addition, MovieNet-TeViS covers 19 diverse movie genres.
More details are presented in the Appendix ~\ref{sup:data details}.

\noindent\textbf{Concreteness measurement.}
The concreteness level of texts has a large influence on visualization difficulty.
To systematically measure the concreteness of texts, we leverage a concreteness database introduced by Brysbaert et al~\cite{Brysbaert2014ConcretenessRF} to calculate average concreteness of words in synopsis and compare with text descriptions of other datasets, i.e., LSMDC, MAD, and CMD and show the results in Tab.~\ref{tab:dataset}. To be specific, Brysbaert \etal ~\cite{Brysbaert2014ConcretenessRF} create a database that ask annotators to assign concreteness ratings from 1 to 5 for 40 thousand English words. The average ratings can evaluate the degree of how concrete a concept denoted by a word is. The larger value means more concrete. For example, the concreteness rating of ``banana'' is 5 while that of ``love'' is 2.07. As shown in the Tab.~\ref{tab:dataset}, our dataset has 2.76 average concreteness ratings while LSMDC and MAD have 2.99. This means that the text synopsis in MovieNet-TeViS is more abstract or higher-level than descriptions in LSMDC and MAD. CMD has 2.60 concreteness score which is slightly lower than ours. This makes sense because CMD use 18 words on average to describe 132-second video clips whereas our MovieNet-TeViS uses 24 words to describe 64-second video segments. As the first trial of a new task, our dataset has appropriate concreteness.

\noindent\textbf{Diversity measurement.} 
Following \cite{wang2019vatex}, we use the number of words, the number of unique n-grams and the number of words with different POS tags to compare diversity of text description or synopsis in LSMDC, MAD, CMD and our dataset. For fair comparison, we randomly sample 10,000 texts from LSMDC, MAD, and CMD datasets. 
We find that our built dataset MovieNet-TeViS has the richest n-grams, nouns, verbs, adjectives and adverbs. 
Due to space limitation, we only show the number of words and the number of unique bi-grams results in Tab.~\ref{tab:dataset}. 
We present the full comparison in the supplementary material.
From Tab.~\ref{tab:dataset}, we observe that CMD is also richer than LSMDC. This supports our observation that LSMDC has caption based descriptions whereas CMD and MovieNet have high-level summaries of movie clips or segments. Our dataset is richer than CMD, which is consistent with what the seconds per word show. 
When looking into our dataset, we find that many text synopses contain dialogues, psychological descriptions, shot languages, etc. Such free text styles are closer to that of our target non-professional video makers. 

\section{Text Synopsis to Video Storyboard Task}
The Text Synopsis to Video Storyboard (\textbf{TeViS}) task aims to retrieve a set of keyframes and order them to visualize the text synopsis. Assume we have a text synopsis $T=\{w_1, w_2, ..., w_n\}$ with $n$ words, the goal of \textbf{TeViS} task is to retrieve $m$ images from large candidate images and order them to visualize the text synopsis. The number of images $m$ is different for each text synopsis $T$. We design two evaluation settings for the \textbf{TeViS} task: i) ordering the shuffled keyframes conditioned on the text, and ii) the task of retrieving and then ordering.

\subsection{Ordering the Shuffled Keyframes}\label{subsec:order}
\noindent\textbf{Task Definition.}
For a given text synopsis and its shuffled ground-truth images, how well can the models order them?
This is a key step for creating a storyboard that needs to consider coherence across frames.
To measure the long-term reasoning capability of models for ordering, we let the models order the ground-truth images for this evaluation. 

\noindent\textbf{Task Evaluation.}
For the ordering task, we are given a text synopsis and its shuffled ground-truth images, the models need to predict their order conditioned on text synopsis. We then can compute $Kendall’s\ \tau$ ~\cite{MirellaLapata2006kendall} metric to report the result.
\begin{equation}
Kendall's\ \tau = 1-\frac{2*\#Inversions}{m*(m-1)/2}
\end{equation}
where $\#Inversions$ are the number of inverse-order pairs, i.e., the number of steps needed to switch to the original order, and $m$ is the number of frames.  $Kendall's\ \tau$ is always between -1 and 1, with 1 representing the full positive order and -1 representing the full inverse order.

\subsection{Retrieve-and-Ordering Keyframes}\label{subsec:overall_task}
\noindent\textbf{Task Definition.}
For a given text synopsis, how well can the models select the relevant images from a large set of candidates and then order them? This task is more practical in real situations. 

\noindent\textbf{Task Evaluation.}
For this evaluation, we are given a text synopsis and a large set of candidate images. The candidate images contain ground-truth images annotated by humans and other negative images which are randomly sampled from other images in the corpus. The number of candidates including ground-truth and negative samples is 500. 

After retrieving top K from the 500 candidates and re-ordering the top K results by different methods, we can calculate $R@K$ and 
$Kendall's\ \tau$ to compare them. However, as some ground-truth images cannot be returned at top K, $Kendall's\ \tau$ is calculated upon the returned ground-truth images at top K only. Consequently, it happens that when fewer correct images are returned at the top, the $Kendall's\ \tau$@K tends to be higher as there are less inversions. This issue can be cured somehow by multiplying R@K and $Kendall's\ \tau$@K:
\begin{equation}
R@K*Kendall's\ \tau@K
\end{equation}
This is used as the final measurement of the retrieve-and-ordering task.

\section{Method}
\begin{figure*}[!t]
    \centering
    \includegraphics[width=0.95\linewidth]{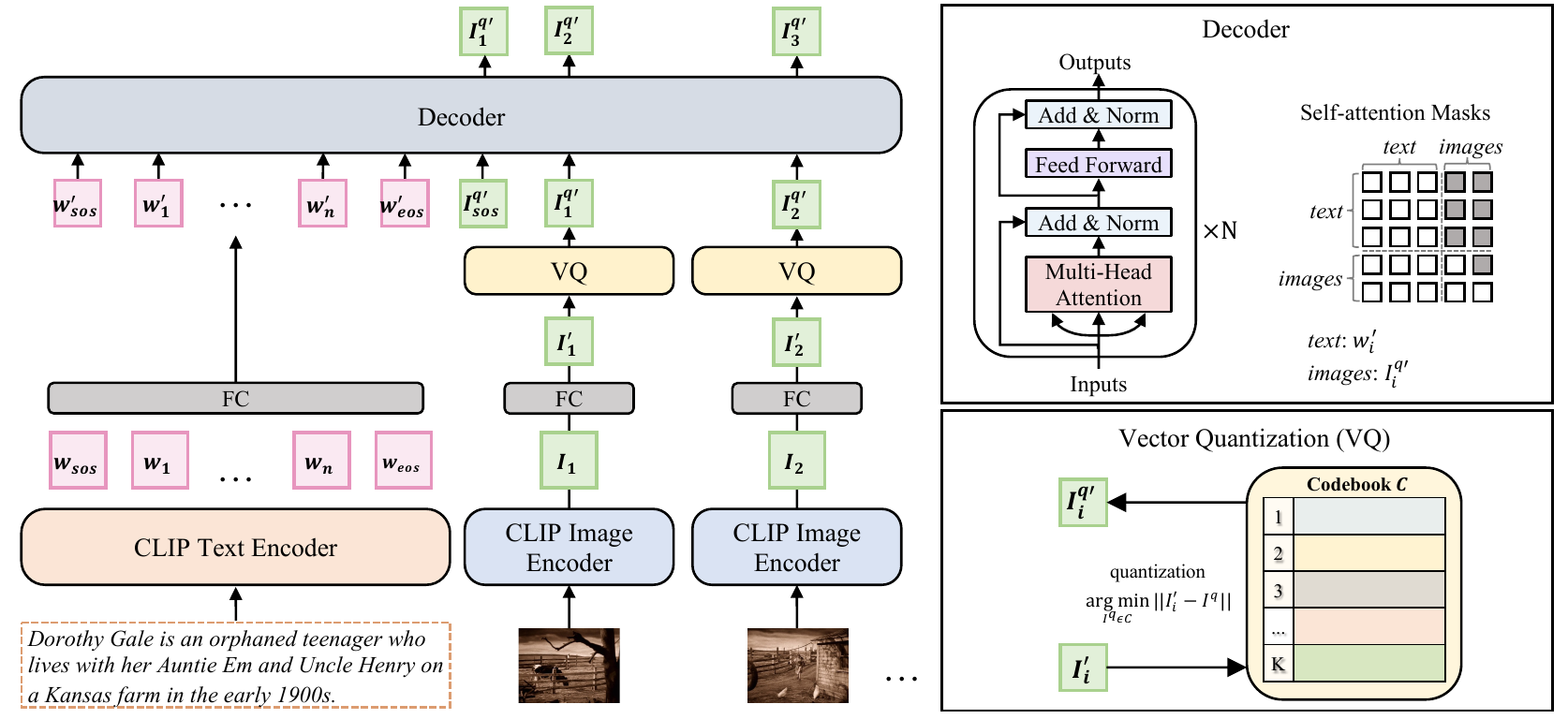}
    \caption{Overview of our ~\modelname~ Model for \textbf{TeViS} task. 
    The text and image extract feature by a fine-tuned CLIP encoder module. 
    In order to enhance the representation ability of images and learn visual semantic features, we used a Vector Quantization(VQ) module to map continuous frame features from the continuous space to the discrete space. Frames are represented by the nearest codebook entry in the codebook $C$ through similarity calculation, yielding a discrete image representation.
    Then the decoder-only transformer will autoregressively generate discrete frame features with text representations as a prefix and can abort the translation by judging $\left \langle eos\right \rangle$ in the inference phase.
    }
    \label{fig:framework}
\end{figure*}

To provide a starting point for tackling the task, we propose a text-to-image retrieval model based on a pre-trained image-text model (i.e., CLIP~\cite{Alec2021CLIP}), and a decoder-only model for ordering images. 
In the retrieval stage, we leverage a pre-trained visual language model (CLIP) to achieve text-to-image retrieval.
(Sec.~\ref{subsec:retrival})
In the first stage of ordering, we utilize a pre-trained visual language model (CLIP) to extract word-level and frame-level representations independently, while simultaneously mapping both to the same semantic space. 
Furthermore, a Vector Quantization module is used to discretize the continuous frame representations.  
In the second stage, a prefix language model is trained over the text token and discrete frame tokens sequence generated by the first stage.
(Sec.~\ref{subsec:model})

\subsection{Text-to-Image Model for Retrieval}\label{subsec:retrival}
\subsubsection{Model}
To establish a connection between textual and visual modalities, we draw inspiration from the extensive employment of large models in multimodal domains.
We leverage a pre-trained image-text model CLIP~\cite{Alec2021CLIP}to conduct text-to-keyframe retrieval.
During training, we randomly sample one frame from the ground-truth keyframe sequence to create a positive image-text pair and frames from other sequences as negatives for a text synopsis.

\subsubsection{Training}
We leverage a contrastive loss to maximize the similarity of matched images and texts while minimizing the similarity of unmatched images and texts, which is: 
\begin{equation}
\begin{aligned}
\mathcal{L}_{i 2 t} &=-\frac{1}{B} \sum_{i=1}^{B} \log \frac{\exp \left(I_{i}^{\top} T_{i} / \tau\right)}{\sum_{j=1}^{B} \exp \left(I_{i}^{\top} T_{j} / \tau\right)}, \\
\mathcal{L}_{t 2 i} &=-\frac{1}{B} \sum_{i=1}^{B} \log \frac{\exp \left(T_{i}^{\top} I_{i} / \tau\right)}{\sum_{j=1}^{B} \exp \left(T_{i}^{\top} I_{j} / \tau\right)}
\end{aligned}
\end{equation}
where $I_i$ and $T_j$ are the normalized embeddings of $i$-th image and $j$-th sentence in a batch of size $B$ and $\tau$ is the temperature. The overall text-image alignment loss $\mathcal{L}_{align}$ is the average of $\mathcal{L}_{i2t}$ and $\mathcal{L}_{t2i}$. 

\subsubsection{Inference}
During inference, we calculate the similarities between the text synopsis and a set of frame candidates and retrieve the top K frames. 

\subsection{Decoder-only Model for Ordering}\label{subsec:model}

\subsubsection{Model}

Inspired by the success of decoder-only framework in Large Language Model,
we propose a \textbf{~\modelname~}model that adopts decoder-only architecture to \textbf{Trans}late Text synopsis to Video Storyboard.
This design can not only sort the candidate images but also handle the variable length problem when creating video storyboards.
In addition, our storyboard dataset exhibits a high degree of visual diversity, making it challenging for the model to learn the language of movie shots.
To overcome the visual diversity problem, we introduce a vector quantization (VQ) method to convert images from continuous visual features into discrete visual tokens.

As illustrated in Fig.~\ref{fig:framework}, 
given synopsis text tokens $\{w_1, w_2, ..., w_n\}$ and the history images $I_{<t}$, our ~\modelname~ will predict the latent feature of the next image $I_t$.

\noindent\textbf{Text encoding.}
We employ a fine-tuned CLIP text encoder to encode the text $w_i$ into $w_i^{\prime}$, along with global text features $w^{\prime}_{sos}$ and $w^{\prime}_{eos}$. All these text features will work as prefix inputs of our decoder.

\noindent\textbf{Image encoding with vector quantization (VQ).}
And we use a fine-tuned CLIP image encoder to encode the frame $I_i$ into $I_i^{\prime}$.
Subsequently, we compute the cosine similarity between each $I_i^{\prime}$ and the discrete features of VQ codebook $C$ and choose the most similar $I_i^{q\prime}$ as the discrete feature of image $I_i^{\prime}$ encoded by VQ.

\noindent\textbf{Transformer decoder.}
Then, given the synopsis text token features $\{w_{sos}^{\prime}, w_1^{\prime}, w_2^{\prime}, ..., w_n^{\prime}, w_{eos}^{\prime}\}$ and the quantized history image features $I^{q\prime}_{<t}$, ~\modelname~ decoder generates a sequence of visual tokens auto-regressively, where each token can be used to retrieve images from the candidate image pool and abort the translation by judging $\left \langle eos\right \rangle$ during inference.
These generated discrete visual tokens can be used to retrieve images through dot-product similarity.
To be more specific, the text prefix and discrete frame tokens are fed into the same transformer decoder without a cross-attention layer, which proves more effective than with cross-attention in our experiments, and use the same position embeddings. We use bi-directional self-attention over text prefixes and use monotonous self-attention on discrete frame tokens.

\subsubsection{Training}
The model is optimized with an NCE loss $\mathcal{L}_{trans}$ and a Vector Quantization loss $\mathcal{L}_{vq}$.

\textbf{NCE loss} in each prediction step with negative images sampled randomly from a mini-batch:
\begin{equation}
\mathcal{L}_{trans} =-\frac{1}{BM} \sum_{i=1}^{B} \sum_{m=1}^{M} \log \frac{\exp \left(I_{i,m}^{\top} I_{i,m}^{'} / \tau\right)}{\sum_{I^{'} \in \mathcal{N}_{i,m} \cup I_{i,m}} \exp \left(I_{i,m}^{\top} I^{'} / \tau\right)}
\end{equation}
where $I_{i,m}$ is the normalized embeddings of the $m$-th image from the $i$-th image sequence from the batch, $\mathcal{N}_{i,m}$ is the normalized embeddings of the negative images sampled from the batch.

\textbf{Vector Quantization loss} for image  feature sequence discretization:
\begin{eqnarray}
\begin{aligned}
I_i^{q} 
&= \ \mathbf{q}(I_i) \
=: \argmin_{I_i^q \in \mathcal{C}}(I_i - I_i^q), \\
\mathcal{L}_{vq} 
&= \sum_{I_i \in \mathcal{N}_{i,m} \cup I_{i,m}} 
\| \text{sg}[I_i] - I_i^{q} \|^2 + \beta \| I_i - \text{sg}[I_i^{q}] \|^2 
\end{aligned}
\end{eqnarray}
where $I_i$ and $I_i^q$ are the i-th frame features and quantized frame tokens, $\mathbf{q}(\cdot)$ is the Vector Quantization module, $\text{sg}[\cdot]$ is the stop-gradient operator, $\beta$ is a commitment loss hyperparameter set to 0.8 in all our experiments.

Thus, the total training objective becomes:
\begin{equation}
\mathcal{L} = \mathcal{L}_{trans} + \mathcal{L}_{vq}
\end{equation}

\subsubsection{Inference}
During inference, the input text is first fed into an encoder block to extract the text representation, and the decoder-only transformer autoregressively generates frame features conditioned on these text features and a $\left \langle sos\right \rangle$ embedding. The generated image feature and the candidate discrete frame embeddings which use the VQ module are compared for similarity, and the top 1 similarity discrete frame embedding is selected as the predicted result. The vector of the top 1 similar image retrieved is then concatenated with the input for the next frame's prediction in the decoder. This top 1 frame is deleted from the candidate discrete frame pool.

\section{Experiments}
We evaluate the performance of proposed methods on the MovieNet-TeViS dataset for the Text Synopsis to Video Storyboard task. We first describe the setup of the experiment(Sec.~\ref{subsec:setup}) and then present the results of both the ordering task(Sec.~\ref{subsec:order_res}) and the retrieve-and-ordering task(Sec.~\ref{subsec:retrival_res}). Finally, we show some qualitative results(Sec.~\ref{subsec:q_res}).

\subsection{Experimental Setup}\label{subsec:setup}
\noindent\textbf{Datasets.}
We randomly divided our MovieNet-TeViS dataset into three separate datasets, consisting of 8035, 1059, and 906 synopsis-keyframes pairs, respectively, for training, validation, and testing. And we ensured that there were no overlapping movies across the subsets.

\noindent\textbf{Implementation Details.}
We utilize CLIP-ViT-B/32 as the backbone in all compared methods. The initial learning rate is set to 1e-6, and we use a linear learning rate scheduler to decay the learning rate linearly after a warm-up stage. The network is optimized by AdamW optimizer, with the weight decay value of 5e-2 and the batch size of 16. 
For vector quantization, we empirically set a codebook of size K as 4096 with 32 dimensions, three layers, and a commitment loss hyperparameter of 0.8 for the $\mathcal{L}_{vq}$.

\subsection{Ordering Task}\label{subsec:order_res}
\noindent\textbf{Compared methods.}\label{subsec:baselines}
In addition to the proposed ~\modelname~ model, we design three strong baselines based on CLIP for ordering,
as shown in Appendix Fig.~\ref{fig:clipbaselines}.
And during inference, our proposed model is able to predict a $\left \langle eos\right \rangle$ symbol to stop, which does not need to know the number of frames in the inference phase. CLIP-based baselines may use the average number.

1) CLIP-Naive: we use CLIP model to calculate the similarity between a text synopsis used as a query and its corresponding keyframes, and then order the keyframes based on the similarity scores.

2) CLIP-Sliding: we first divide the sentences into several segments as a group of queries where the number of segments is equal to the number of its corresponding keyframes.  We then use a sliding window to use each segment to retrieve the most similar keyframes in turn. Once a keyframe is chosen, this keyframe will be removed from the candidates.

3) CLIP-Cumulative: we first divide the sentences into several segments as CLIP-Sliding. However, when doing retrieval, we accumulate each segment and retrieve the most similar keyframes, which consider more context. For example, to retrieve the second keyframe, we use the first two segments as the query. We also remove the keyframes from the candidates once they are chosen in the previous step.

4) Neural Storyboard~\cite{chen2019storyboard} and CNSI~\cite{Ravi2018ShowMA}:we train the state-of-the-art story-to-image models ~\cite{chen2019storyboard, Ravi2018ShowMA} on our dataset.
In the CNSI model, we used the same way as CLIP-Cumulative to adapt the models for one-to-many text-image retrieval.

5) Re-Ranking: we replace our decoder with a standard transformer block and predict the indexes of input frames. 

\begin{table}[!t]
\centering
\caption{Results of ordering task. Our method ~\modelname~ achieves the best overall performance, though still leaving much room for improvement compared to human capabilities. $[s-e]$ under Kendall's $\tau$ denotes sequence length from $s$ to $e$.}

\begin{tabular}{lccc}
\toprule
\multirow{2}*{Method}  & \multicolumn{3}{c}{Kendall's\ $\tau$$\uparrow$} \\
\cmidrule{2-4}
& Over-All & [3-5] & [6-11] \\
\midrule
Neural Storyboard~\cite{chen2019storyboard}  & 0.163 & 0.214 & 0.047 \\
CNSI~\cite{Ravi2018ShowMA} &0.182	&0.234	&0.064\\
Re-Ranking       & 0.192 & 0.252 & 0.058 \\
\midrule
CLIP-Naive    & 0.183 & 0.248 & 0.036 \\
CLIP-Sliding  & 0.230 & 0.278 & 0.123\\
CLIP-Cumulative  & 0.244 & 0.291 & 0.139\\
\midrule
~\modelname~(ours)  & \textbf{0.367} & \textbf{0.407} & \textbf{0.278}\\
\midrule
Human  &  0.821 & 0.860 & 0.734\\
\bottomrule
\end{tabular}
\label{tab:result}
\end{table}

\begin{table}[!t]
\centering
\caption{Ablation study of image Vector Quantization (VQ) method and the combination ways of text and images.}
\begin{tabular}{ccccc}
\toprule
\multicolumn{2}{c}{Method} &  \multicolumn{3}{c}{Kendall's\ $\tau$$\uparrow$} \\
\cmidrule{3-5}
Image-VQ. &Text & Over-All & [3-5] & [6-11] \\
\midrule
\ding{55} &Cross &0.224 &0.291	&0.071  \\
\ding{55} &Prefix &0.230 &0.287 &0.102   \\
\ding{51} &Cross &0.268 &0.327  &0.133    \\
\ding{51} &Prefix &\textbf{0.367} &\textbf{0.407} &\textbf{0.278}   \\
\bottomrule
\end{tabular}
\label{tab:abl_vq_text}
\end{table}

\begin{figure*}
    \centering
    \includegraphics[width=\linewidth]{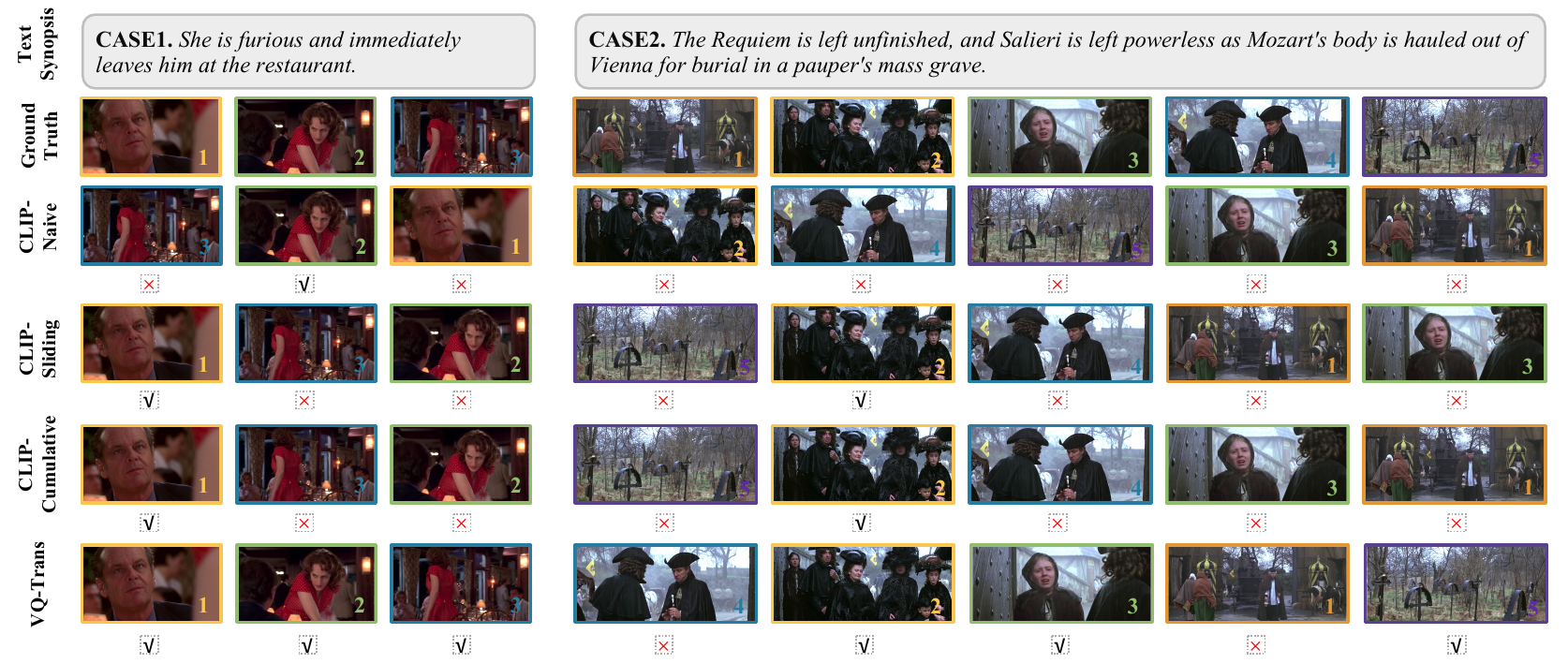}
    \caption{Qualitative examples of different models for the ordering task on our Movie-TeViS dataset. 
    }
    \label{fig:qualitive_results}
\end{figure*} 

\noindent\textbf{Main Results.}
We conducted several experiments to verify the effect of different methods on the text synopsis to video storyboard task, and results are shown in Tab.~\ref{tab:result}. 
The CLIP-Naive method achieves the poorest performance due to the lack of sequence modeling. The CLIP-Sliding and CLIP-Cumulative methods outperform the CLIP-Naive method, proving to be more effective ways of applying CLIP, thanks to the ability to segment semantic information of the text and thus able to model sequences. 
We also compare with two SOTA methods of story-to-image task~\cite{chen2019storyboard, Ravi2018ShowMA}. The two methods perform significantly worse than CLIP-Sliding and our ~\modelname~. It indicates that CLIP-Sliding is a simple but effective baseline. 
We carry out a comparison with a re-ranking manner in Tab.~\ref{tab:result}.
Our ~\modelname~ method achieves significantly better performance, because it can better decouple semantic representation and ordering, and thus requires fewer data, demonstrating the ability of sequence generation models to learn long-term information in keyframe sequences. And more importantly, the proposed feature prediction model enables an extension to generative models in the future.

In addition, a human study was conducted to make a better assessment of the task. We invited participants to reorder the shuffled keyframe sequences. The performance of humans is presented in Tab.~\ref{tab:result}. Humans achieve much better performance than our best model, suggesting there is high potential for improvement.

\noindent\textbf{Ablations.}
We conduct a ablation study of image vector quantization (Image-VQ) and the combination ways of using text. The results in Tab.~\ref{tab:abl_vq_text} show that by applying Image-VQ, the performance is significantly improved in all metrics. This indicates that Image-VQ is effective to reduce the vision space to learn. When Image-VQ is enable, using text as a prefix significantly outperforms using text to cross-attend images in all metrics. This indicates that our proposed decoder-only method is more effective than the traditionally widely-used encoder-decoder model.

\subsection{Retrieve-and-Ordering Task}\label{subsec:retrival_res}

\begin{table}[!t]
\centering
\caption{Text-to-image retrieval 
performance on MovieNet-TeViS. We compare CLIP model with and without fine-tuning.}

\begin{tabular}{ccccc}
\toprule
Fine-tuning & R@1$\uparrow$ & R@5$\uparrow$ & R@10$\uparrow$ & R@30$\uparrow$ \\
\midrule 
\ding{55}  & 5.73 & 19.72 & 28.98 & 46.90   \\
\ding{51}  & \textbf{7.62} & \textbf{25.99} & \textbf{38.18} &\textbf{58.19}  \\
\bottomrule
\end{tabular}
\label{tab:retrieval}
\end{table}

\noindent\textbf{Retrieval Results.}
We first evaluate the performance of text-to-image retrieval. We compare CLIP models~\cite{Alec2021CLIP} with and without fine-tuning on our MovieNet-TeViS dataset. 
As shown in Tab.~\ref{tab:retrieval}, although CLIP shows reasonable performance by zero-shot, the fine-tuned CLIP model can achieve significantly better performance in all metrics.

\noindent\textbf{Retrieve-and-Ordering Results.}
We report the result of the Retrieve-and-Ordering Task in Tab.~\ref{tab:result_overall}. 
Initially, a pool of image candidates was retrieved from 500 images (including positive and negative examples) by fine-tuning the CLIP model.
Subsequently, diverse models were employed to retrieve the top K images from this candidate set. In this process, $Kendall's \ \tau$ was used to evaluate the ranking of ground truth among the top K images. Finally, the performance evaluation of the task was based on $R@K * Kendall's \ \tau@K$.
It indicates that the method of ~\modelname~ achieves the best performance while Neural Storyboard, CNSI, and Re-Ranking are the overall poorest, with CLIP related methods in the middle. This result is basically consistent with the experimental results of the ordering task, suggesting that the difference in performance primarily stems from the difference in methods' ability of ordering.

\subsection{Qualitative Results}\label{subsec:q_res}
In addition to the quantitative results, we further carry out a case study on how well our proposed methods perform in the TeViS task. 
As Fig.~\ref{fig:qualitive_results} shows, for the given text synopsis, 
our proposed ~\modelname~ method performs the best and can correctly order all three frames in Case 1, as well as the No.2, 3, and 5 images in Case 2.  
CLIP-Naive takes the synopsis as the whole to encode and thus it actually considers relevance only without any order information. It performs worst as expected. Our proposed CLIP-Sliding and CLIP-Cumulative address the limitation because text synopsis is split into several text fragments and the ordering of keyframes depends on the ordering of text fragments. 
In this case, the text fragments are well aligned with ground-truth keyframes from human's perspective, 
but it is still difficult for CLIP-Sliding and CLIP-Cumulative in ordering these cases.
Our proposed pre-training and transformer based model can correctly order the keyframes, which shows the advantages in learning the visual language for storyboard creation.

\begin{table}[htp]
\centering
\caption{Results of Retrieve-and-Ordering Task. Our method ~\modelname~ outperforms other CLIP-based methods. 
We use $R@K*Kendall's \ \tau@K$ as the final measurement.
}
\begin{tabular}{lcc}
\toprule
Method & K=20 & K=30 \\
\midrule
Neural Storyboard~\cite{chen2019storyboard}  & 0.116 & 0.107 \\
CNSI~\cite{Ravi2018ShowMA} & 0.114 & 0.126\\
Re-Ranking       & 0.124 & 0.121 \\
\midrule
CLIP-Naive      & 0.203 & 0.078\\
CLIP-Sliding    & 0.202 & 0.131\\
CLIP-Cumulative & 0.219 & 0.138\\
\midrule
~\modelname~    & \textbf{0.300} & \textbf{0.219}\\ 
\bottomrule
\end{tabular}
\label{tab:result_overall}
\end{table}
\vspace{-0.3cm}
\section{Conclusion}

In this paper, we introduce a novel \textbf{TeViS} task (\textbf{Te}xt synopsis to \textbf{Vi}deo \textbf{S}toryboard), which aims to retrieve an ordered sequence of images to visualize the text synopsis.
We also construct a \textbf{MovieNet-TeViS} dataset to support it. 
To align the diverse text synopsis with keyframes, we utilize a pre-trained Image-Text model to overcome this challenge. 
We propose a decoder-only model called \textbf{~\modelname~} which translates text synopsis to keyframe sequence. 
We also propose a VQ module on Movies frames to discrete the continuous Movies frames representations.
Ablation studies verify the effectiveness of our proposed model. Both quantitative and qualitative results show our method is better than other baselines.

\begin{acks}
This work was supported by the Fundamental Research Funds for the Central Universities, and the Research Funds of Renmin University of China (21XNLG28), National Natural Science Foundation of China (No. 62276268) and Bilibili Inc. - Research on Artificial Intelligence Assisted Storyboarding (No. 2020K20880).
We acknowledge the anonymous reviewers for their helpful comments. 
\end{acks}

\bibliographystyle{ACM-Reference-Format}
\balance
\bibliography{ref}

\clearpage
\appendix       

\setcounter{figure}{6} 
\setcounter{table}{5} 
\renewcommand{\thefigure}{\thesection.\arabic{figure}} 
\renewcommand{\thetable}{\thesection.\arabic{table}} 

\section{Limitation and Social Impact}\label{sup:limit social}
We propose a new dataset that consists of text synopses and corresponding keyframes. During the annotation progress, we maintained a professional visual language. Our new task is based on this dataset that aims to create video storyboards for text synopses. Thus we hope our work can help amateurs when creating their videos by providing references.
However, the movie styles that we consider now are not very comprehensive, such as the design of camera angles, and how the camera moves, etc. These will be considered in our future works.

\section{MovieNet-TeViS Dataset Details}\label{sup:data details}
\setcounter{table}{0}
\setcounter{figure}{0}

\begin{table*}[!t]
\centering
\caption{Diversity statistics for our dataset and other datasets.}
\begin{tabular}{lccccccc}
\toprule
             & \multicolumn{3}{c}{\textbf{\#unique n-grams}} &\multicolumn{4}{c}{\textbf{\#unique words (POS tags)}}\\
\cmidrule(lr){2-4} \cmidrule(lr){5-8}   
\textbf{Dataset} & 1-gram & 2-gram & 3-gram & noun & verb & adjective & adverb \\
\midrule
$LSMDC_{10k}$ ~\cite{rohrbach2017lsmdc} & 10,137 & 44,010 & 67,231 & 29,989 & 15,085 & 4,486 & 3,686 \\
$MAD_{10k}$~\cite{soldan2022mad} & 14,215 & 59,352 & 89,686 & 38,336 & 17,219 & 5,567 & 4,260 \\
$CMD_{10k}$ ~\cite{bain2020cmd} & 17,566 & 83,202 & 121,992 & 53,070 & 24,152 & 7,181 & 3,151 \\
$MovieNet-TeViS$ & \textbf{24,325} & \textbf{134,524} & \textbf{248,021} & \textbf{108,832} & \textbf{97,235} & \textbf{21,946} & \textbf{30,559} \\
\bottomrule
\end{tabular}
\label{tb:sup text rich}
\end{table*}

\subsection{Keyframe sequence length and text length}
\begin{figure}[htp]
   \centering
   \subfloat[Description of the keyframe length.]
   {\begin{minipage}[b]{0.23\textwidth}
       \centering       \includegraphics[width=\linewidth]{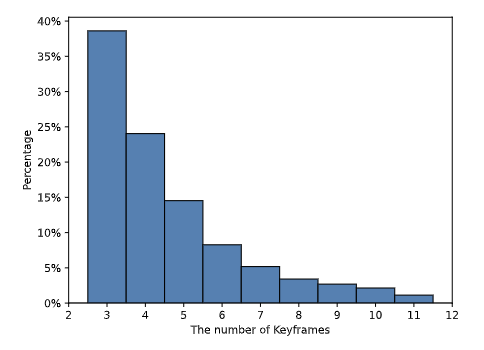}
       \label{fig:keyframe}
   \end{minipage}}
   \hfill
   \subfloat[Description of the sentence length.]{
   \begin{minipage}[b]{0.23\textwidth}
       \centering       \includegraphics[width=\linewidth]{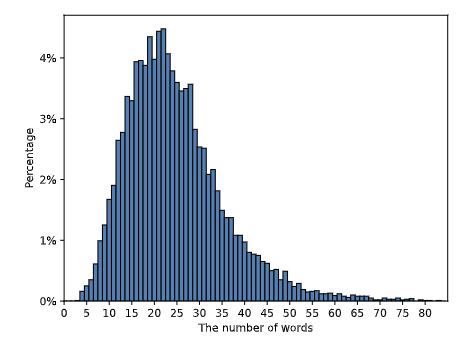}
       \label{fig:desc_token}
   \end{minipage}}
   \hfill
   \subfloat[Distribution of movie genres.]{
   \begin{minipage}[b]{0.23\textwidth}
       \centering       \includegraphics[width=\linewidth]{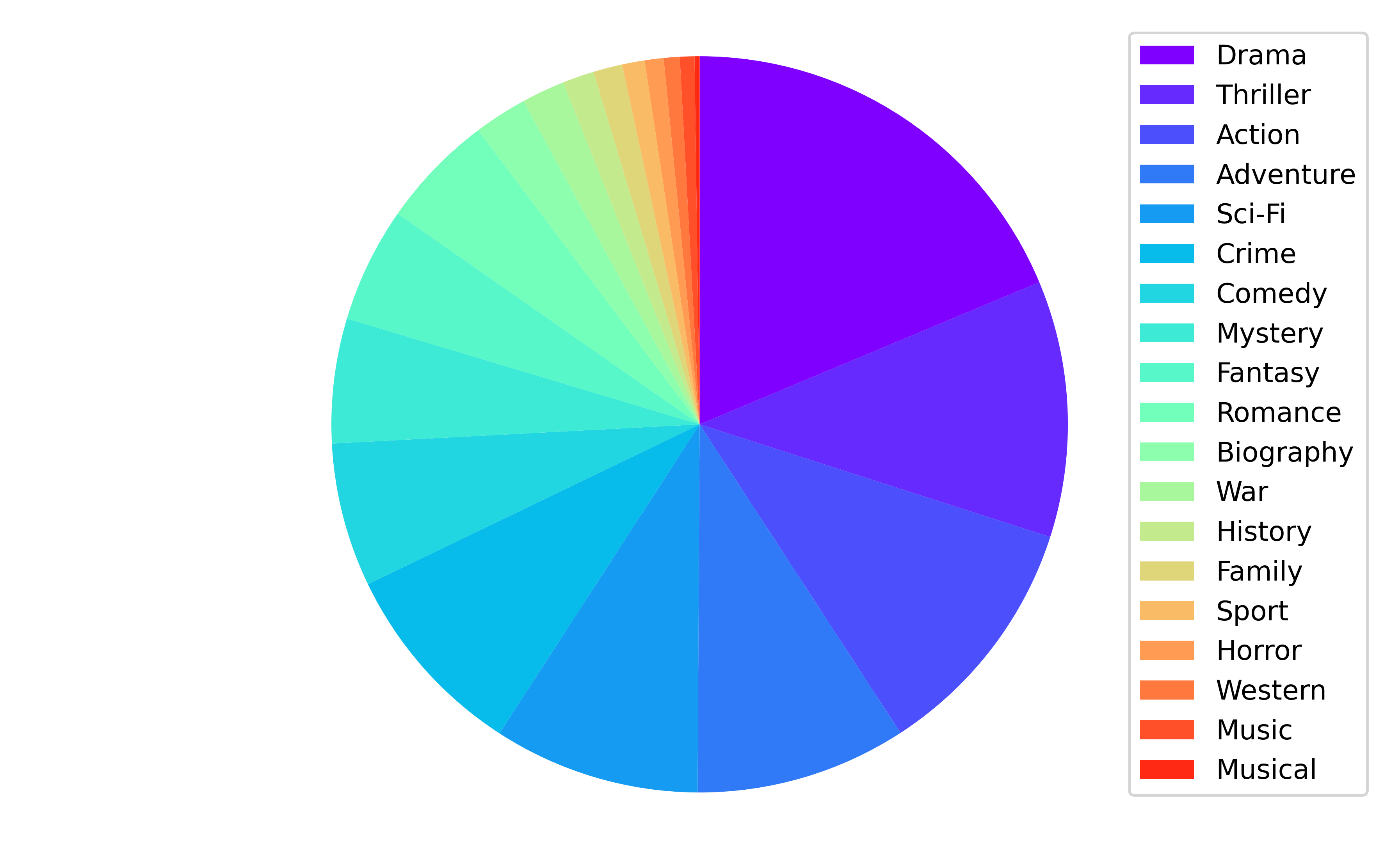}
       \label{fig:movie_genres}
   \end{minipage}}
\caption{More detailed statistics of MovieNet-TeViS dataset.}
\label{fig:dataset_statistics}
\end{figure}

We select 2,949 paragraph-segment pairs from MovieNet~\cite{huang2020movienet} for labeling by removing some pairs which are offensive. We ask annotators to label these data and obtain 10,000 pairs of synopses and keyframe sequences. There are 45,584 keyframes in total. As Fig.\ref{fig:keyframe} shows, the number of keyframes is from 3 to 11 and about 60\% storyboards contain 3 or 4 keyframes. Fig.\ref{fig:desc_token} shows the distribution of the number of words in a text synopsis. The peak is around 20 and most of the sentences contain less than 80 words. In addition, as Fig.\ref{fig:movie_genres} shows, MovieNet-TeViS contains movies with diverse genres. There is no dominant category. The top 1 genre occupies about 20\% and the top three genres occupy only about 40\%.

\subsection{Semantic Richness}
The text synopses in our dataset are rich in script types, including dialogue, scene descriptions, mental activities, etc. We compare the semantic richness of our dataset with previous movie datasets~\cite{rohrbach2017lsmdc,soldan2022mad,bain2020cmd} in terms of the average unique n-grams and part-of-speech (POS) tags. For a fair comparison, we randomly selected 10K examples from LSMDC~\cite{rohrbach2017lsmdc}, CMD~\cite{bain2020cmd}, MAD~\cite{soldan2022mad} dataset. As Tab.~\ref{tb:sup text rich} shows, our MovieNet-TeViS dataset is higher in both unique n-grams and POS tags, which can further reflect the semantic richness of our dataset.

\section{Implementation details}
\begin{figure}[H]
    \centering \includegraphics[width=0.95\columnwidth]{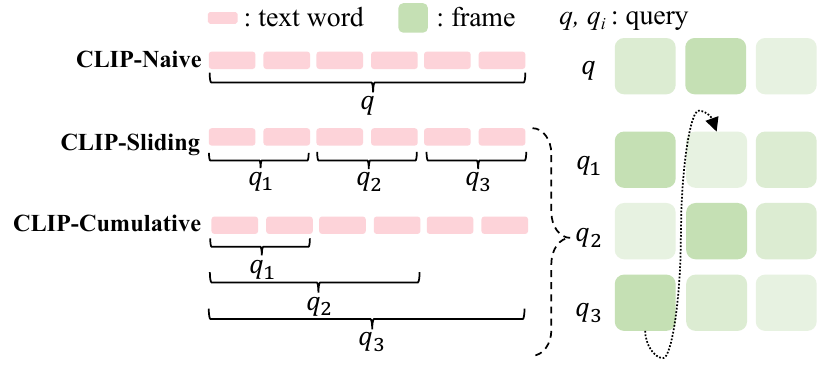}
    \caption{Illustration of additional baseline models for ordering. }
    \label{fig:clipbaselines}
\end{figure}

We design three strong baselines based on CLIP for ordering,
as shown in Fig.~\ref{fig:clipbaselines}.
\subsection{Potential improvement with alternate CLIP models}
\parskip=0.1em
\begin{enumerate}
[itemsep=0.1em,parsep=0em,topsep=0em,partopsep=0em]
    \item CLIP-Dynamic: It may be challenging to adjust the number of segments as it needs to match the number of keyframes. We have implemented a dynamic segmentation approach for dividing the synopsis into segments (considering complexity, we adopt 10,000 divisions). While the number of segments remains the same across all divisions, the cutting positions vary. We use bipartite graph matching to calculate the optimal matching between segments and keyframes. Based on this optimal matching result, we report the final metrics.
    \item CLIP-Contextual: Our CLIP-Cumulative method is equivalent to CLIP-Contextual with a greedy strategy. We implement CLIP-Contextual with a beam search strategy and show the results in Tab. ~\ref{tab:clip_more_baseline}. As the table shows, the gap between CLIP-Contextual with a beam search strategy and CLIP-Cumulative is relatively small.
\end{enumerate}

\begin{table}[H]
\centering
\caption{Comparing results of our methods with story-to-image baselines and CLIP-based methods in ordering subtask.}
\begin{tabular}{lccc}
\toprule
\multirow{2}*{Method}  & \multicolumn{3}{c}{Kendall's\ $\tau$$\uparrow$} \\
\cmidrule{2-4}
& Over-All & [3-5] & [6-11] \\
\midrule
CLIP-Sliding	&0.230	&0.278	&0.123\\
CLIP-Cumulative	&0.244	&0.291	&0.139\\
\midrule
CLIP-Dynamic	&0.218	&0.250	&0.146\\
CLIP-Contextual	&0.241	&0.287	&0.138\\
\midrule
~\modelname~(ours)  & \textbf{0.367} & \textbf{0.407} & \textbf{0.278}\\
\bottomrule
\end{tabular}
\label{tab:clip_more_baseline}
\end{table}

\subsection{Limitations of Vector Quantization (VQ) and contrastive loss}
We have conducted an analysis to investigate the impact of codebook dimension and size on the ordering task, as shown in Tab. ~\ref{tab:more_vq_test}. 
Based on the results, we selected a codebook size of 4,096 and a dimension of 32. 

\begin{table}[H]
\centering
\caption{Exploring the impact of codebook dimensions and sizes.}
\begin{tabular}{ccc}
\toprule
dim & codebook\_size & Kendall's $\tau$ \\
\midrule
32 & 1024 & 0.349\\
32 & 4096 & 0.367\\
32 & 8192 & 0.338\\
64 & 1024 & 0.319\\
64 & 4096 & 0.353\\
64 & 8192 & 0.355\\
128 & 1024 & 0.348\\
128 & 4096 & 0.344\\
128 & 8192 & 0.346\\
512 & 1024 & 0.344\\
512 & 4096 & 0.335\\
512 & 8192 & 0.348\\
\bottomrule
\end{tabular}
\label{tab:more_vq_test}
\end{table}

In addition, we investigated alternative variants of vanilla Vector Quantization (vanilla-VQ) in our study, namely Multi-Stage Vector Quantization (MS-VQ) \cite{lee2022autoregressive}, Soft Vector Quantization (Soft-VQ) \cite{guo2021soft}, and Hierarchical Vector Quantization (Hi-VQ) \cite{Ali2019Advances}, as recommended. a) We utilized a three-stage approach to model residual VQ for MS-VQ. b) We employed SoftMax membership calculation to model soft VQ for Soft-VQ. c) We implemented a two-layer hierarchical approach for the Hi-VQ variant.
The results are presented in Tab. ~\ref{tab:more_vq}. Experimental results show that different variants of VQ tend to perform better at lower dimensions. The variants did not exhibit the expected superiority over vanilla-VQ. Moreover, we find that the training process of MS-VQ, including the loss curve and validation curve, was remarkably stable compared to other implemented versions. MS-VQ may be promising to achieve better performance. 

\begin{table}[H]
\centering
\caption{Comparison of different vector quantization variants.}
\begin{tabular}{ccccc}
\toprule
dim, size & Vanilla-VQ & MS-VQ & Soft-VQ & Hi-VQ \\
\midrule
32, 4096 & 0.367 & 0.360 & 0.250 & 0.298 \\
64, 8192 & 0.355 & 0.354 & 0.224 & 0.278 \\
128, 1024 & 0.348 & 0.349 & 0.230 & 0.286 \\
512, 8192 & 0.348 & 0.320 & 0.235 & 0.213 \\
\bottomrule
\end{tabular}
\label{tab:more_vq}
\end{table}

\subsection{Balancing loss functions and sensitivity to hyper-parameters}
For instance, we introduce a coefficient $\lambda$ to linearly combine the two losses in Formula 6: $L = L\_trans + \lambda * L\_vq$. Then we investigate the impact of various values on the final performance and present preliminary results in Tab. ~\ref{tab:loss}. The results indicate that our current setting, specifically $\lambda = 1$, yields the best performance.

\begin{table}[H]
\centering
\caption{Performance with different coefficients of the loss function.}
\begin{tabular}{lccc}
\toprule
\multirow{2}*{Method}  & \multicolumn{3}{c}{Kendall's\ $\tau$$\uparrow$} \\
\cmidrule{2-4}
& Over-All & [3-5] & [6-11] \\
\midrule
$\lambda$ = 0.1 & 0.345 & 0.384 & 0.258 \\
$\lambda$ = 1 (ours)  & 0.367 & 0.407 & 0.278 \\
$\lambda$ = 10  & 0.364 & 0.396 & 0.290 \\
\bottomrule
\end{tabular}
\label{tab:loss}
\end{table}

\end{document}